\documentclass[11pt,a4paper]{article}
\usepackage[hyperref]{emnlp2020}
\usepackage{times}
\usepackage{latexsym}

\usepackage[utf8]{inputenc}
\usepackage{xcolor}
\usepackage[ruled,vlined]{algorithm2e}
\usepackage{times}
\usepackage{latexsym}

\usepackage{graphicx}
\usepackage{multirow}
\usepackage{booktabs}
\usepackage{todonotes}
\usepackage{subfigure}

\usepackage{url}

\usepackage{bm}
\usepackage{amsfonts, amsmath}
\usepackage{amsthm}
\usepackage{bbm}
\usepackage{url}
\usepackage{graphicx}
\usepackage{microtype}

\newcommand{\e}[1]{{\small $#1$}}

\makeatletter
\def\adl@drawiv#1#2#3{%
        \hskip.5\tabcolsep
        \xleaders#3{#2.5\@tempdimb #1{1}#2.5\@tempdimb}%
                #2\z@ plus1fil minus1fil\relax
        \hskip.5\tabcolsep}
\newcommand{\cdashlinelr}[1]{%
  \noalign{\vskip\aboverulesep
           \global\let\@dashdrawstore\adl@draw
           \global\let\adl@draw\adl@drawiv}
  \cdashline{#1}
  \noalign{\global\let\adl@draw\@dashdrawstore
           \vskip\belowrulesep}}
\makeatother

\aclfinalcopy 

\title{MCMH: Learning Multi-Chain Multi-Hop Rules\\ for Knowledge Graph Reasoning}

\author{Lu Zhang$^{~\clubsuit}$ \quad Mo Yu$^{~\heartsuit}$ \quad Tian Gao$^{~\heartsuit}$ \quad Yue Yu$^{~\clubsuit}$
\\
   $^\clubsuit$ Lehigh University \quad 
   $^\heartsuit$ IBM Research \quad 
   \\
{\small
    \texttt{luz319@lehigh.edu} \quad
    \texttt{\{yum, tgao\}@us.ibm.com}\quad 
 \texttt{yuy214@lehigh.edu}\quad
  }}

\date{}

\begin{document}
\maketitle
\begin{abstract}
Multi-hop reasoning approaches over knowledge graphs infer a missing relationship between entities with a multi-hop rule, which corresponds to a chain of relationships.
We extend existing works to consider a generalized form of multi-hop rules, where each rule is a set of relation chains. 
To learn such generalized rules efficiently, we propose a two-step approach that first selects a small set of relation chains as a rule and then evaluates the confidence of the target relationship by jointly scoring the selected chains.
A game-theoretical framework is proposed to this end to simultaneously optimize the rule selection and prediction steps.
Empirical results show that our multi-chain multi-hop (MCMH) rules result in superior results compared to the standard single-chain approaches, justifying both our formulation  of  generalized rules  and the effectiveness of the proposed learning framework.

\end{abstract}

\section{Introduction}
\label{sec:intro}

Knowledge graphs (KGs) represent knowledge of the world as relationships between entities, i.e., triples with the form \emph{(subject, predicate, object)} \cite{Bollacker2008,suchanek2007semantic,Vrandei2014,10.1007/978-3-540-76298-0_52,Carlson2010}. Such knowledge resource provides clean and structured evidence for many downstream applications such as question answering.
KGs are usually constructed by human experts, which is time-consuming and leads to highly incomplete graphs \cite{min-etal-2013-distant}. Therefore automatic KG completion \cite{nickel_three-way_2011,bordes2013translating,yang2014embedding,chen2018variational,reasontensor,lao2011random} is proposed to infer a missing link of relationship $r$ between a head entity $h$ and a tail entity $t$.

\begin{figure}
\includegraphics[width=1.0\linewidth,height=2.4in]{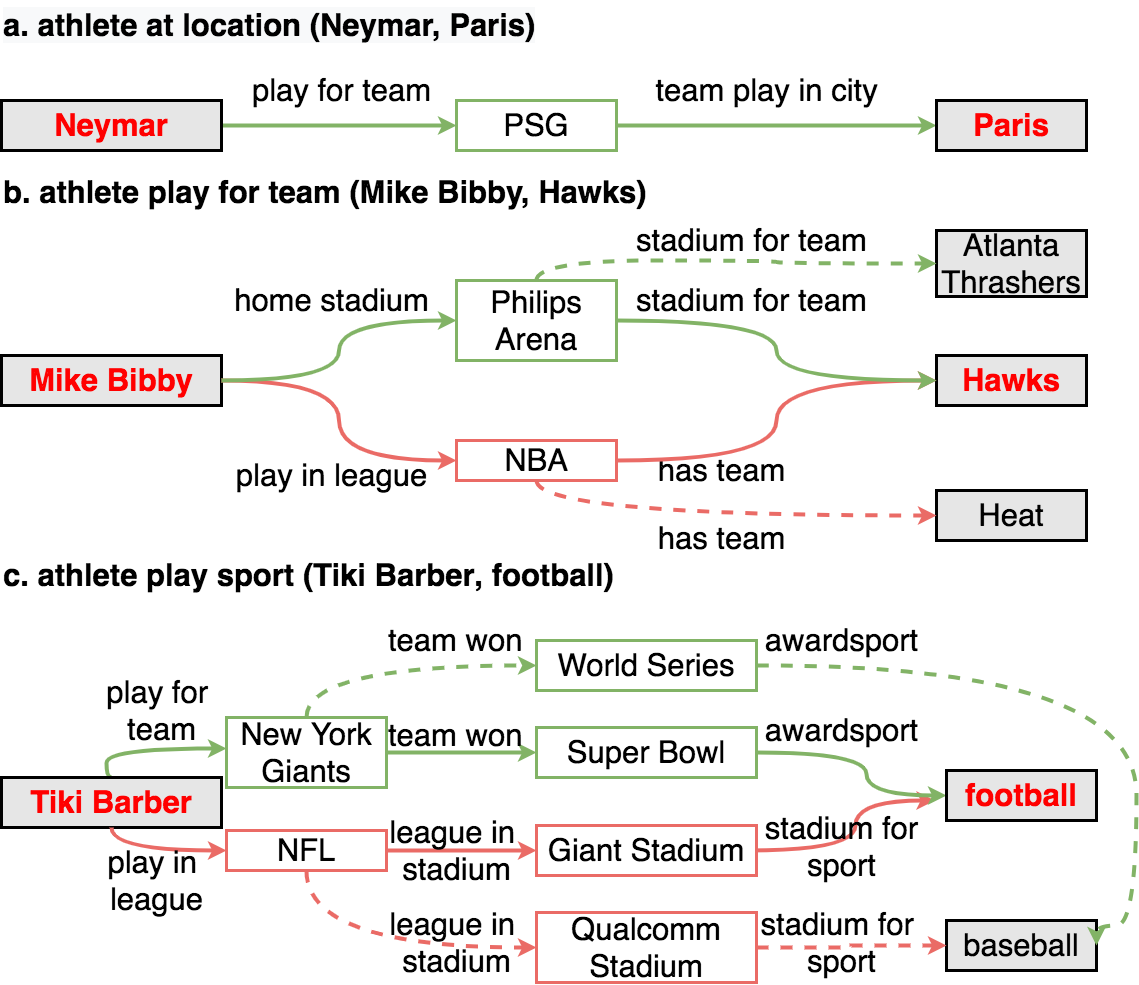}
\caption{\small{Examples of reasoning with multiple paths. (a) A standard multi-hop example. The target can be sufficiently inferred with one chain. (b) An example that requires a rule as the conjunction of two chains (the stadium hosts two teams but only one from NBA). (c) An example where multiple chains cannot sufficiently infer the target but improves its confidence.} \label{fig:example}}
\vspace{-0.2in}
\end{figure}

Existing KG completion work mainly makes use of two types of information: 1) co-occurrence of entities and relations and 2) deducible reasoning paths of tuples. \textbf{KG embeddings} encode entities and relations, the first type of information, together into continuous vector space with low-rank tensor approximations~\citep{bordes2013translating,dettmers2017convolutional,lin2015learning,NeelakantanRM15,ShiandWeninger,trouillon2016complex,wang2014knowledge,Xie2016,yang2014embedding}.

Ours approach utilizes the second type of information, reasoning path of tuples that can be deduced to the target tuple~\citep{lao2010relational,xiong2017deeppath,das2016chains,das2017go}.
Here a reasoning path starts with the head entity $h$ and ends with the tail entity \e{t}: \e{h \overset{r_1}{\rightarrow} e_1  \overset{r_k}{\rightarrow} e_k \overset{r_N}{\rightarrow} t}, where \e{r_1 \wedge ... \wedge r_N} forms a relation chain that infers the existence of $r$. Therefore these methods are also referred as \textbf{multi-hop reasoning over KGs}, which learns a multi-hop chain as a rule to deduce the target $r$. An example of such a chain is given in Figure~\ref{fig:example}a to infer whether an athlete plays in an location. Multi-hop reasoning approaches can usually utilize richer evidence and self-justifiable in terms of  reasoning path rules used in the predictions, making the prediction of missing relations more interpretable.

Despite  advantages and  success of the multi-hop reasoning approach \cite{multihoplin,xiong2017deeppath,das2017go,shen2018reinforcewalk:,chen2018variational,VariationalRea}, a target relationship may not be perfectly inferred from a single relation chain. There could exist multiple weak relation chains that correlate with the target relation.
Figure~\ref{fig:example} gives examples of such cases. 
These multiple chains could be leveraged in following ways: (1) the reasoning process naturally relies on the logic conjunction of multiple chains (Figure~\ref{fig:example}b); (2) more commonly, there are instances for which none of the chains is accurate, but aggregating multiple pieces of evidence improves the confidence (Figure~\ref{fig:example}c), as also observed in the case-based study works~\cite{aamodt1994cbr,das2020cbr}.
Inspired by these observations, we propose the concept of  \textbf{multi-chain multi-hop rule set}. 
Here, instead of treating each single multi-hop chain as a rule, we learn rules consisting of a small set of multi-hop chains. Therefore the inference of target relationships becomes a joint scoring of such  a set of chains. {We  treat each set of chains as one rule and, since different query pairs can follow different rules, together we have  a set of rules to reason each relation.}

Learning the generalized multi-hop rule set is a combinatorial search problem. 
We address this challenge with a game-theoretic approach inspired by~\cite{lei2016rationalizing,carton2018extractive,yu2019rethinking}.
Our approach consists of two steps: (1) selecting a generalized multi-hop rule set by employing a Multi-Layer Perceptron (MLP) over the candidate chains;  (2) reasoning with the generalized rule set, which uses another MLP to model the conditional probability of the target relationship given the selected relation chains.
The nonlinearity of MLP as reasoner provides the potential to model the logic conjunction among the selected chains in the rule set.

We demonstrate the advantage of our method on KG completion tasks in FB15K-237 and NELL-995. Our method outperforms existing single-chain approaches, showing that our defined generalized rules are necessary for many reasoning tasks.

\section{Backgrounds}

\noindent\textbf{Problem Formulation}
We aim to infer missing relationships between two given entities, such as \texttt{athleteAtLocation} between \texttt{Neymar} and \texttt{Paris}, given their other connections in the knowledge graph. 
Formally, we are given a knowledge graph $\mathcal{G}$, consisting of a set of triplets ${O = \{(h, r, t)\}}$, where $r$ is a relation edge defined in $\mathcal{G}$,
$h$ is a head entity, and $t$ is the tail entity. 
The task is to identify the relation ${\hat{r}}$ between a set of query entity ${\hat{h}}$ and ${\hat{t}}$. For evaluation, we have ground truth labels indicating whether each pair ${(\hat{h},\hat{t})}$ has the relationship ${\hat{r}}$ or not.

For a given query $(\hat{h_i}, \hat{r},\hat{t_i})$, the $i$-th sample in  $\hat{r}$, we extract a set of relation chains \e{\mathcal{R}=\{{\bm R}_n\}_{n=1}^{N}=\{(\hat{h},r_n^1,t_n^1),(t_n^1,r_n^2,t_n^2),\cdots(t_n^{m-1},r_n^m,\hat{t})\}_{n=1}^N} from the original KB $\mathcal{G}$.  Each chain is a set of connected relations between $\hat{h}$ and $\hat{t}$ in $\mathcal{G}$. 
The proposed \textbf{multi-chain multi-hop rule set} is a set of rules, each consisting of multiple relation chains $\mathcal{S} \subset\mathcal{R}$ with size $d=|\mathcal{S}|$.
In the experiments, we represent each relation chain \e{{\bm R}_n} with only relation names.
Our task is to find such $\mathcal{S}$ 
for a target relation $\hat{r}$ over each query pair $\hat{h}_i$ and $\hat{t}_i$, and estimate the confidence  $P(\hat{r} |\mathcal{S})$. Note that $\mathcal{S}$ and $\mathcal{R}$ depend  on query sample  $(\hat{h_i}, \hat{r},\hat{t_i})$
but for notation simplicity we omit $i$ and $\hat{r}$ from $\mathcal{S}_i^{\hat{r}}$ and $\mathcal{R}_i^{\hat{r}}$.

\noindent\textbf{Relation Chains Extraction}
\label{ssec:chain_extraction}
To obtain the set of candidate relation chains  $\mathcal{R}$ for a target relation $\hat{r}$, we take the following extraction steps. First, we extract a fixed hop $k$ sub-graph from the original KB. Each sub-graph starts with an entity $\hat{h}$ with relation $\hat{r}$, ends with an entity $\hat{t}$, and satisfies that $(\hat{h}, \hat{r}, \hat{t}) \in \mathcal{G}$.
The sub-graph consists of a list of $m$-hop paths connecting the two ends, where $1\leq m\leq k$. Each of the $m$-hop paths has the form \e{(\hat{h},r^1,t^1), (t^1,r^2,t^2), \cdots (t^{m-1},r^{m},\hat{t})}. We call \e{r^1\rightarrow r^2 \cdots \rightarrow r^{m}} a candidate relation chain  \e{\bm R}. High $k$ values can result in an intractable number of chains while low $k$ values may not have sufficient coverage. Here we extract chains with length up to $k=3$, and for $\hat{r}$ with a large number of chains ($|\mathcal{R}|\geq10^4$), we  filter out extracted chains with a set threshold (proportional to count of relation chains) in the positive training data for that relation.

\section{A Game-Theoretic Approach for MCMH Rule Learning}

\begin{figure*}
\center
\includegraphics[width=\textwidth,height=8cm]{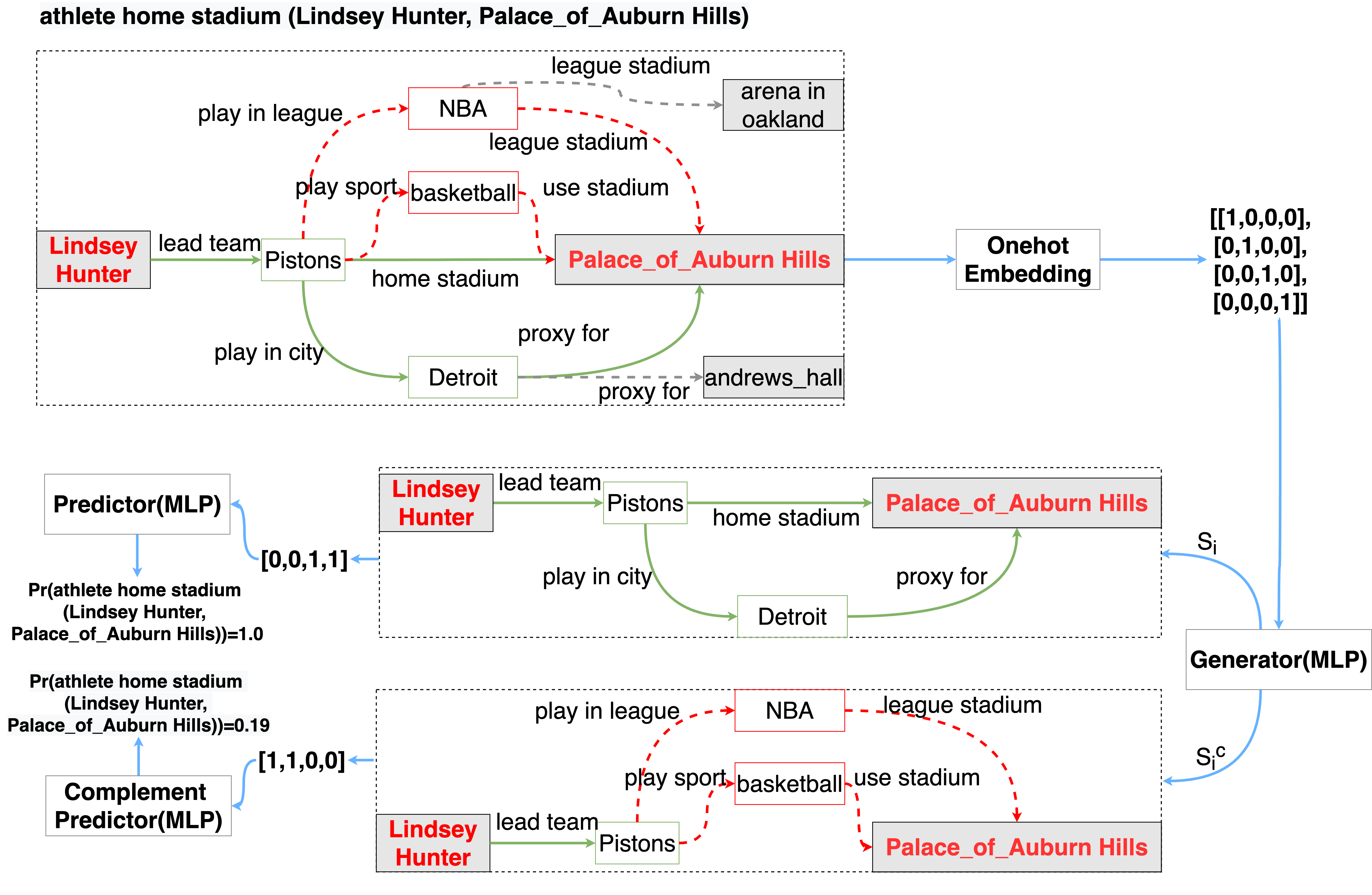}

\caption{\small{An example workflow of our model, with $|\mathcal{R}_i|=4$. The generator selects the first two chains as the ``critical information'' for prediction:  \e{\mathcal{S}_i=\{\texttt{LeadTeam}\rightarrow\texttt{HomeStadium},\texttt{LeadTeam}\rightarrow\texttt{PlayinCity}\rightarrow\texttt{Proxyfor}\}} with complement \e{\mathcal{S}^c_i=\{\texttt{LeadTeam}\rightarrow\texttt{PlayinLeague}\rightarrow\texttt{LeagueStadium},\texttt{LeadTeam}\rightarrow\texttt{PlaySport}\rightarrow\texttt{UseStadium}\}}. In the prediction phase, the predictor $\mathcal{S}_i$ is encoded as $\mathbf{v}_{{\mathcal{S}_i}}=[0,0,1,1]$ and estimates probability of \texttt{athleteHomeStadium} being true as $100\%$. The complement predictor $\mathcal{S}^c_i$ is encoded as $\mathbf{v}_{{\mathcal{S}^c_i}}=[1,1,0,0]$ and estimates the probablity as $19\%$. 
} \label{fig:example2}}

\end{figure*}

\noindent\textbf{A Three-Player Game for Rule Learning}
Finding a set of chains as the rule is a combinatorial search problem in $\mathcal{R}$. For example, given an input of 1,000 chains between a training entity pair, the selection of a set-rule of 4 chains corresponds to a search space of 10$^{12}$.
Hence, we propose a game-theoretic approximation to learn to generate predictive chains and reduce the learning complexity. Our method is inspired by the line of rationalization works~\cite{carton2018extractive,yu2019rethinking}. 
Specifically, our input is a set of chains $\mathcal{R}_i \subset \mathcal{R}$ 
for relation $\hat{r}$ and each training sample $(\hat{h}_i,\hat{r},\hat{t}_i)$. Our method consists of three submodels: (1) a \emph{rule set generator} that selects the set of chains $\mathcal{S}_i$ as a rule, (2) a \emph{reasoner} that predicts the probability of $\hat{r}_i$ based on $\mathcal{S}_i$, and (3) a \emph{complement predictor} that predicts the probability of $\hat{r}$ based on $\mathcal{S}_i^c= \mathcal{R}_i \setminus \mathcal{S}_i$.

During training, the \emph{predictor} and the \emph{complement predictor} aim to minimize the cross-entropy loss for predicting the existence of $\hat{r}$. While the \emph{generator} is optimized to make the \emph{predictor} perform well, while decreasing the \emph{complement predictor}'s accuracy. In other words, the \emph{generator} plays a cooperative game with the \emph{predictor} to make the selected rule set $\mathcal{S}_i$ be useful for inferring the target relationship $\hat{r}$. At the same time it plays an adversarial game with the \emph{complement predictor} to ensure that no critical information is left, i.e., to ensure the comprehensiveness of the selected  $\mathcal{S}_i$. 
An example of the workflow is given in Figure~\ref{fig:example2}.

\color{black}
\noindent\textbf{Predictors} The predictor estimates probability of $\hat{r}$ being true conditioned on $\mathcal{S}_i$, denoted as $\hat{p}(\hat{r}|\mathcal{S}_i)$. The complement predictor estimates probability of $\hat{r}$ conditioned on $\mathcal{S}_i^c$, denoted as $\hat{p}^c(\hat{r}|\mathcal{S}_i^c)$. The two models are optimized as follows: 
\begin{equation}
    \small
    \begin{aligned}
    & \mathcal{L}_p = \min_{\hat{p}} -H(p(\hat{r} | \mathcal{S}_i); \hat{p}(\hat{r} | \mathcal{S}_i)), \\
    & \mathcal{L}_c = \min_{\hat{p}^c} -H(p(\hat{r} | \mathcal{S}_i^c); \hat{p}^c(\hat{r} | \mathcal{S}_i^c)),
    \end{aligned}
    \label{eq:loss_pred}
\end{equation}
where ${H(p;q)}$ denotes the cross entropy between ${p}$ and ${q}$, and ${p(\cdot | \cdot)}$ denotes the empirical distribution. 

We encode the inputs ${\mathcal{S}_i}$ and ${ \mathcal{S}_i^c}$  as binary vectors $\mathbf{v}_{{\mathcal{S}_i}}$ and $\mathbf{v}_{{\mathcal{S}^c_i}}$, respectively\footnote{Our method could use KG embedding as inputs like previous works~\cite{xiong2017deeppath,das2017go}. It may weakens the interpretability of the reasoning model as they are smoothed representations, but can potentially improve the performance for cases with smaller training data. We leave the investigation to future work.}, which are both of dimension $\vert \mathcal{R}_i \vert$, with each dimension corresponding to one relation chain in the candidate set $\mathcal{R}_i$. The $j$-th component of $\mathbf{v}_{{\mathcal{S}_i}}$ is set to $1$ if and only if the $j$-th chain is selected in ${\mathcal{S}_i}$, i.e., \e{\bm{R}}$_j\in\mathcal{S}_i$, and similarly for $\mathbf{v}_{{\mathcal{S}^c_i}}$. The input vectors are fed into a 3-layer MLP to predict whether $\hat{r}$ holds for $(\hat{h}_i, \hat{t}_i)$. 

\noindent\textbf{Generator} The generator extracts $\mathcal{S}_i$ from the input chain set $\mathcal{R}_i$. 
This function, denoted as 
$g:\mathcal{R}_i\rightarrow \mathcal{S}_i$, is optimized with:
\begin{equation}
\small
    \min_{\bm g(\cdot)} \mathcal{L}_p - \mathcal{L}_c + \lambda_s \mathcal{L}_s,
    \label{eq:loss_gen}
\end{equation}
where \e{\mathcal{L}_p} and \e{\mathcal{L}_c} are the losses of the \emph{predictor} and the \emph{complement predictor}, respectively.
\e{\mathcal{L}_s} is a sparsity loss which aims to constrain the number of chains to be select to a desired size $d$:
\begin{equation}
    \small
    \mathcal{L}_s = \max\{(\vert \mathcal{S}_i \vert - d)/\vert \mathcal{R}_i \vert, 0 \}.
    \label{eq:Ls_Lh}
\end{equation}

Since the \emph{generator} makes a hard decision for selection of $\mathcal{S}_i$, the losses \e{\mathcal{L}_p} and \e{\mathcal{L}_c} are generally not differentiable. Hence, we utilize the policy gradient~\cite{Williams:1992} reinforcement learning algorithm to optimize the generator. To have bounded rewards, we use the predictors' accuracy instead of the loss values  \e{\mathcal{L}_p} and \e{\mathcal{L}_c}. %
The generator is also modeled with a MLP that is of the same architecture as the predictor. 
The output is a $\vert \mathcal{R}_i\vert \times 2$ vector which represents the probabilities that each chain would be selected  into $\mathcal{S}_i$ and $\mathcal{S}_i^c$ .

\noindent\textbf{Rule selection during inference}
During inference, to have a fixed number ($d$) of selection, for each instance, we select the top-$d$  chains  according to the probability predicted by the generator.

\begingroup
\setlength{\tabcolsep}{4pt}
{\tiny\begin{table}[t]
\center
\addtolength{\tabcolsep}{0.1pt}
{\small\begin{tabular}{c c c c c} 
\toprule
\bf Dataset& \bf \#Entity& \bf \#Relation &\bf \#Triples &\bf  \#Tasks\\
\midrule
FB15K-237&14,505 & 237 & 310,116 &10\\
NELL-995 &75,492 &200 &154,213&10\\
\bottomrule
\end{tabular}}
\caption{Statistics of the Datasets.}
\label{table:stats}
\end{table}}
\endgroup

\begin{table*}[t]
\center
\addtolength{\tabcolsep}{0.05pt}
{\small

\begin{tabular}{l r r  l rr }
\toprule
\multicolumn{3}{c}{\bf FB15K-237} & \multicolumn{3}{c}{\bf NELL-995}\\
 \cmidrule(lr){1-3} 
 \cmidrule(lr){4-6}
\bf Relation & \bf \#Chains & \bf \#Chains per Sample & \bf Relation & \bf \#Chains & \bf \#Chains per Sample  \\
\midrule
teamSports & 115&5.1&athletePlaysForTeam & 852&20.9\\
birthPlace & 285 &62.5&athletePlaysInLeague & 568&6.2\\
filmWrittenBy & 153 &65.9&athleteHomeStadium & 174&5.2 \\
filmDirector& 132 &37.5&athletePlaysSport & 143&3.3\\
filmLanguage & 3,380&82.2&orgHeadquaterCity & 2,467&16.2\\
tvLanguage & 1,614 &55.2&orgHiredPerson & 4,717&20.7\\
capitalOf & 2,634&117.1 &bornLocation & 974&23.8\\
orgFounded & 3,728 &102.9&personLeadsOrg & 3,347&20.3\\
musicianOrigin & 6,784 &158.2&teamPlaySports & 228&6.3 \\
personNationality&365&49.0&worksFor & 4,840&21.6\\
\bottomrule
\end{tabular}}
\caption{Number of chains extracted for each relation. We show both the total number of different chains for each relation, and the average number of chains that can be extracted per instance.} 
\label{tab:num_of_chains}
\end{table*}

{\footnotesize\begin{table*}[t]
\centering
\addtolength{\tabcolsep}{0.1pt}
{\footnotesize\begin{tabular}{c l c c c c c cc} 
\toprule
& \multirow{2}{*}{\bf Relation}&\bf  Single-Chain &  \multicolumn{2}{c}{\bf Ours} & \multicolumn{2}{c}{\bf Ours (-conj)} & \multirow{2}{*}{\bf DeepPath} &\multirow{2}{*}{ \bf MINERVA}\\
&  &\bf Baseline & \bf $\bm d$=2& \bf $\bm d$=5 &\bf $\bm d$=2 &\bf $\bm d$=5 & &\\
\midrule
\multirow{11}{*}{\rotatebox{90}{\bf NELL-995}}
& athletePlaysForTeam & 0.872&0.940${^*}$& \bf 0.947${^*}$&0.900&0.897&0.750 &0.824\\
& athletePlaysInLeague &0.962 &0.977${^*}$ & \bf 0.981${^*}$&0.957&0.975&0.960& 0.970 \\
& athleteHomeStadium & 0.892&\bf0.896&0.895&0.856&0.854&0.890& 0.895\\
& athletePlaysSport & 0.916&0.978${^*}$&0.982${^*}$&0.932&0.978&0.957 &\bf0.985\\
& teamPlaySports & 0.728 &0.769 &0.782&0.669&0.771 &0.738&\bf0.846\\
& orgHeadquarterCity &0.957&0.932&0.907&\bf 0.962&0.903&0.790&0.946\\
& worksFor & 0.794 &0.842${^*}$ & \bf0.849${^*}$&0.811&0.842&0.711 &0.825\\
& bornLocation &0.823&\bf 0.902${^*}$&0.850${^*}$&0.874&0.872&0.757&0.793\\
& personLeadsOrg & 0.833&0.832&0.813&0.832& 0.822& 0.795&\bf0.851\\
& orgHiredPerson & 0.833&0.825&0.814&0.837&\bf 0.855&0.742&0.851\\
\cline{2-9}
& \emph{Average} &\it 0.861&\textbf{\emph{0.890}}&\it 0.882&\it 0.863&\it 0.877&\it 0.809&\it 0.879
\\
\midrule
\midrule
\multirow{11}{*}{\rotatebox{90}{\bf FB15K-237}}
& teamSports &0.740&0.739&0.769${^*}$&0.758&0.765&\bf0.955&-\\
& birthPlace	&0.463	&0.505${^*}$	&\bf 0.566${^*}$	&0.443	&0.512	&0.531&-\\
& filmDirector&0.303&0.368&0.411${^*}$&0.363&0.413&\bf 0.441&-\\
& filmWrittenBy 	&0.498&0.516${^*}$	&\bf0.553${^*}$	&0.507	&0.518	&0.457&-\\
& filmLanguage&0.632&0.665${^*}$&\bf0.678${^*}$&0.667&0.675&0.670&-\\
& tvLanguage&\bf0.975&0.962&0.957&0.957&0.956&0.969&-\\
& capitalOf &0.648&0.795&\bf0.825${^*}$&0.820&0.786&0.783&-\\
& orgFounded &0.465&0.407&\bf0.490${^*}$&0.431&0.485&0.309&-\\
& musicianOrigin & 0.376&0.408${^*}$&\bf0.516${^*}$&0.390&0.476&0.514& -\\
& personNationality &0.713&0.806${^*}$&\bf 0.828${^*}$&0.703&0.760&0.823&-\\
\cline{2-9}
& \emph{Average} &\it 0.581&\it 0.617	&\textbf{\emph{0.659}}	&\it 0.604	&\it 0.635	&\it 0.645&-\\
\bottomrule
\end{tabular}}
\caption{Overall Results (MAP) on NELL-995 and FB15K-237. $^*$ highlights the cases where our MLP model outperforms the baseline with statistical significance (p-value$<$0.01 in t-test).}
\label{tab:res_nell}
\end{table*}}

\section{Empirical Evaluation}
We evaluate our model with MCMH rules on two datasets, FB15K-237~\cite{toutanova-etal-2015-representing} and NELL-995~\cite{xiong2017deeppath}.
We follow the existing setting of treating each target relationship as a separate task and training and evaluating relationship-specific reasoning models, and use the standard data splits~\cite{xiong2017deeppath}. 
Table~\ref{table:stats} summarizes statistics of two datasets. For each target relation in the datasets, we extract candidate chain set $\mathcal{R}$ following Section~\ref{ssec:chain_extraction}. 
Table~\ref{tab:num_of_chains} shows the number of extracted chains for each relation.
We compare with previous works in the same setting, DeepPath~\cite{xiong2017deeppath} and MINERVA~\cite{das2017go}. They both are single-chain methods, i.e., they learn a reasoning model to find a single multi-hop chain for the inference.

\noindent\textbf{Overall results}
Table~\ref{tab:res_nell}  shows our method with double chains and five chains outperforms the single-chain baseline ($d=1$ in our model) by clear margins on both datasets, demonstrating the advantage of our generalized rules compared to the single-chain rules studied in the existing works. 
Moreover, our generalized rule learning method, when  setting $d=5$, outperforms existing baselines on both datasets.
For some relations (such as the \texttt{teamSports} relation), our method performs worse compared to the previous works. It is likely because the relation has less training data while previous works use pre-trained KG embeddings to alleviate the problem.

\noindent\textbf{Effects of numbers of chains in one rule ($\bm d$)}
The required numbers of chains differ from different datasets: on NELL-995, using double- relation chain with $d=2$ achieves slightly better performance compared to setting $d=5$, while on FB15K-237 there is a clear advantage with  $d=5$ relation chains.
This observation shows that on FB15K-237
a relation generally requires more chains as evidence to improve the confidence of prediction. Moreover, since  a conjunction rule usually does not span over 5 chains, for many FB15K-237 test tuples the evidence is not sufficient for making the decision, therefore adding more chains can enhance the confidence thus improve results significantly.

\noindent\textbf{Choices of $\bm d$} The average number of chains (i.e., the number of chains that connect the specific entity pair) is 13.8 for NELL-995 and 63.3 for FB15K-237. Therefore  selecting $d$=5 chains is a significant  portion of the whole input space. Moreover, MAP of our model using all candidate chains is 0.671 for FB15K-237 and 0.892 for NELL-995, which are close to that of $d$=5 (the detail performance for each relation is shown in Appendix \ref{app:upperbound}). From the above observations, selecting $d$=5 chains is sufficient for the KB completion task.
Also, the logic conjunction between $d$=2 chains or among 5 chains is more likely to be human-interpretable compared to the selection of large numbers of chains.
Figure \ref{fig:curve} of Appendix \ref{app:upperbound} shows MAP versus the number of selected chains $d$ for two representative relations, showing that the performance of our model converges after $d$=5.

\noindent\textbf{Effects of MLP versus linear predictors} 
Finally we study the impact of the two different ways that our generalized rules contribute to the improved results, namely \emph{modeling logic conjunctions} and \emph{enhancing confidence of multiple weak rules}, as discussed in Section~\ref{sec:intro}.
To this end, we replace the MLP predictors with linear models. The rationale is that the linear model is less effective in capturing conjunctions among inputs, so improvements from linear models over the single-chain baseline are more likely due to the enhanced confidence, rather than finding a conjunctive rule.
We denote this model as \textbf{Ours (-conj)} and show its results in Table \ref{tab:res_nell}. It is observed that the \textbf{Ours (-conj)} model outperforms the baseline, but is generally not as good as the MLP model.    Hence most of the relations mainly benefit from the case of confidence enhancement.
However, the results also highlight a few relations with a notable performance gap, e.g., \texttt{athletePlaysForTeam}, indicating that multiple conjunctions are also important to KB completion tasks.

\section{Conclusion}
We propose a new approach of multi-chain multi-hop rule learning for knowledge graph completion tasks. First, we formalize the concept of multi-hop rule sets with multiple relation chains from knowledge graphs. Second, we propose a game-theoretical learning approach to efficiently select predictive relation chains for a query relation. Our formulation and learning method demonstrate advantages on two benchmark datasets over existing single-chain based approaches. For future work, we plan to investigate rules beyond chains, as well as integrate KG embeddings into our framework. 

\section*{Acknowledgments}
L. Zhang and Y. Yu are supported by the National Science Foundation under award DMS 1753031.

\bibliography{emnlp2020}
\bibliographystyle{acl_natbib}

\newpage
\appendix

\onecolumn

\section{Hyper-parameters and Reproducibility Checklist}
\label{app:checklist}
\paragraph{Implementation dependencies libraries} 
Preprocess: networkx 2.4.
Model: Pytorch 1.4.0,  cuda10.2.

\paragraph{Computing infrastructure} The experiments run on servers with Intel(R) Xeon(R) CPU E5-2650 v4 and Nvidia GPUs (can be one of Tesla P100, V100, GTX 1070,or K80). The allocated RAM is 150G. GPU memory is 8G.
	
\paragraph{Model description}
There are 3 parts in our model, predictor, complement predictor, generator.
\begin{itemize}
\item MLP: Each part employs 3 linear layers with ReLU as activation. The dimension of each layer is half of that in the previous layer. 
\item Linear: The generator has the same structure as MLP, but the predictor and complement predictor have only one linear layer.
\end{itemize}

\paragraph{Average runtime for each approach}
\begin{itemize}
\item MLP: The training time varies for each task, ranging from 8 hours to 50 hours. The main factor in the time variance is the size of the combinatorial action space. 
\item Linear: Training time ranges from 4 hours to 30 hours. 
\end{itemize}

\paragraph{Number of model parameters}
The trainable parameter number of our model is task-specific, because the rules number varies for different relation tasks.
For a task with $D$ numbers of rules, the number of parameters of our MLP model is: 
\begin{align*}
  g(D) &= 3(D\times\frac{D}{2} + \frac{D}{2}\times\frac{D}{4}+\frac{D}{4}\times2)= \frac{15D^2}{8}+\frac{3D}{2}. \\
\end{align*}
For instance, in the task of \texttt{personNationality}, $D$ = 365. The number of parameters for each model is:
\begin{itemize}
\item MLP: 250,347
\item Linear: 84,913
\end{itemize}

\paragraph{Corresponding validation performance for each reported test result }	
The validation results of NELL-995 are listed in  Table~\ref{tab:res_dev}. 
{\footnotesize\begin{table*}[h]
\centering
\addtolength{\tabcolsep}{0.1pt}
{\footnotesize\begin{tabular}{c l c c c c c cc} 
\toprule
& \multirow{2}{*}{\bf Relation}&\bf  Single-Chain &  \multicolumn{2}{c}{\bf Ours} & \multicolumn{2}{c}{\bf Ours (-conj)} & \multirow{2}{*}{\bf DeepPath} &\multirow{2}{*}{ \bf MINERVA}\\
&  &\bf Baseline & \bf $\bm d$=2& \bf $\bm d$=5 &\bf $\bm d$=2 &\bf $\bm d$=5 & &\\
\midrule
\multirow{11}{*}{\rotatebox{90}{\bf NELL-995}}
& athletePlaysForTeam &	0.946&	0.964&	0.962&	0.954&	0.955&	0.750&	0.824\\
 &athletePlaysInLeague &	0.963&	0.965&	0.971&	0.955&	0.967&	0.960&	0.970\\
 &athleteHomeStadium &	0.918&	0.931&	0.945&	0.936&	0.922&	0.890&	0.895\\
& athletePlaysSport &	0.942&	0.955&	0.960&	0.934&	0.959&	0.957&	0.985\\
& teamPlaySports &	0.837&	0.830&	0.825&	0.771&	0.830&	0.738&	0.846\\
& orgHeadquarterCity &	0.963&	0.961&	0.959&	0.944&	0.916&	0.709&	0.946\\
& worksFor &	0.902&	0.953&	0.913&	0.938&	0.913&	0.711&	0.825\\
& bornLocation &	0.955&	0.939&	0.950&	0.930&	0.946&	0.757&	0.793\\
& personLeadsOrg &	0.984&	0.966&	0.981&	0.9571&	0.983&	0.795&	0.851\\
 &orgHiredPerson &	0.893&	0.890&	0.886&	0.881&	0.867&	0.742&	0.851\\
 \cline{2-9}
&average&	0.930&	0.935&	0.935&	0.920&	0.926&	0.809&	0.879\\
\bottomrule
\end{tabular}}
\caption{Overall results (MAP) on validation set of NELL-995.} 
\label{tab:res_dev}
\end{table*}}
\paragraph{Explanation of evaluation metrics used}
In our experiment, we use Mean Average Precision (MAP) \cite{Zhang2009} as the evaluation metric. 

\paragraph{Hyper-parameters } We do not conduct extensive hyper-parameter tuning. In all tests we set learning rate of Adam as $0.001$ and batch size as 20. Embedding dimension is the number of rules for each relation task. The weight for sparsity loss is set as 1.0.

\paragraph{Data preprocess }
The statistics of original datasets are shown in Table~\ref{table:stats}.
For the training set, we do the downsampling on the negative samples. We split the training and dev sets with the ratio of 0.8. 

\section{Results with All Chains}
The idea in our paper is reasoning with more than one chains could improve KB completion performance, since they  contain more information. So we perform experiments with $d$=all and show the results in Table~\ref{tab:res_all}. In these experiments there is no generator. All chains between the given query $(\hat{h}, \hat{r},\hat{t})$ are taken as the input of the predictor. From intuition, with more evidence a higher MAP is generally expected. We therefore use these results as a reference upperbound of our method. \footnote{Precisely, this result could not show the real upperbound of reasoning task with more than one chains. This is due to (1) the capacity of the MLP models may not be sufficient to capture the conjunction among all chains; (2) the reported numbers are affected by the generalizability of models and randomness of the data.}
\label{app:upperbound}
\begin{table}[h]
\center
\addtolength{\tabcolsep}{0.05pt}
{\small
\begin{tabular}{l r  l r }
\toprule
\multicolumn{2}{c}{\bf FB15K-237} & \multicolumn{2}{c}{\bf NELL-995}\\
 \cmidrule(lr){1-2} 
 \cmidrule(lr){3-4}
\bf Relation & \bf $\bm d$=all  & \bf Relation & \bf $\bm d$=all  \\
\midrule
teamSports & 0.791 &athletePlaysForTeam & 0.946\\
birthPlace & 0.577 &athletePlaysInLeague & 0.970\\
filmWrittenBy & 0.579 &athleteHomeStadium & 0.864 \\
filmDirector& 0.420 &athletePlaysSport & 0.977\\
filmLanguage & 0.696&orgHeadquaterCity & 0.935\\
tvLanguage & 0.960 &orgHiredPerson & 0.851\\
capitalOf & 0.817 &bornLocation & 0.828\\
orgFounded & 0.508 &personLeadsOrg & 0.836\\
musicianOrigin & 0.527 &teamPlaySports & 0.839 \\
personNationality&0.834&worksFor & 0.869\\
\hline
Average &0.671 &Average &0.892 \\
\bottomrule
\end{tabular}}
\caption{MAP Results of our predictor with all chains ($d$=$\infty$).} 
\label{tab:res_all}
\end{table}

\begin{figure}[h]
\center
\includegraphics[width=0.678\linewidth]{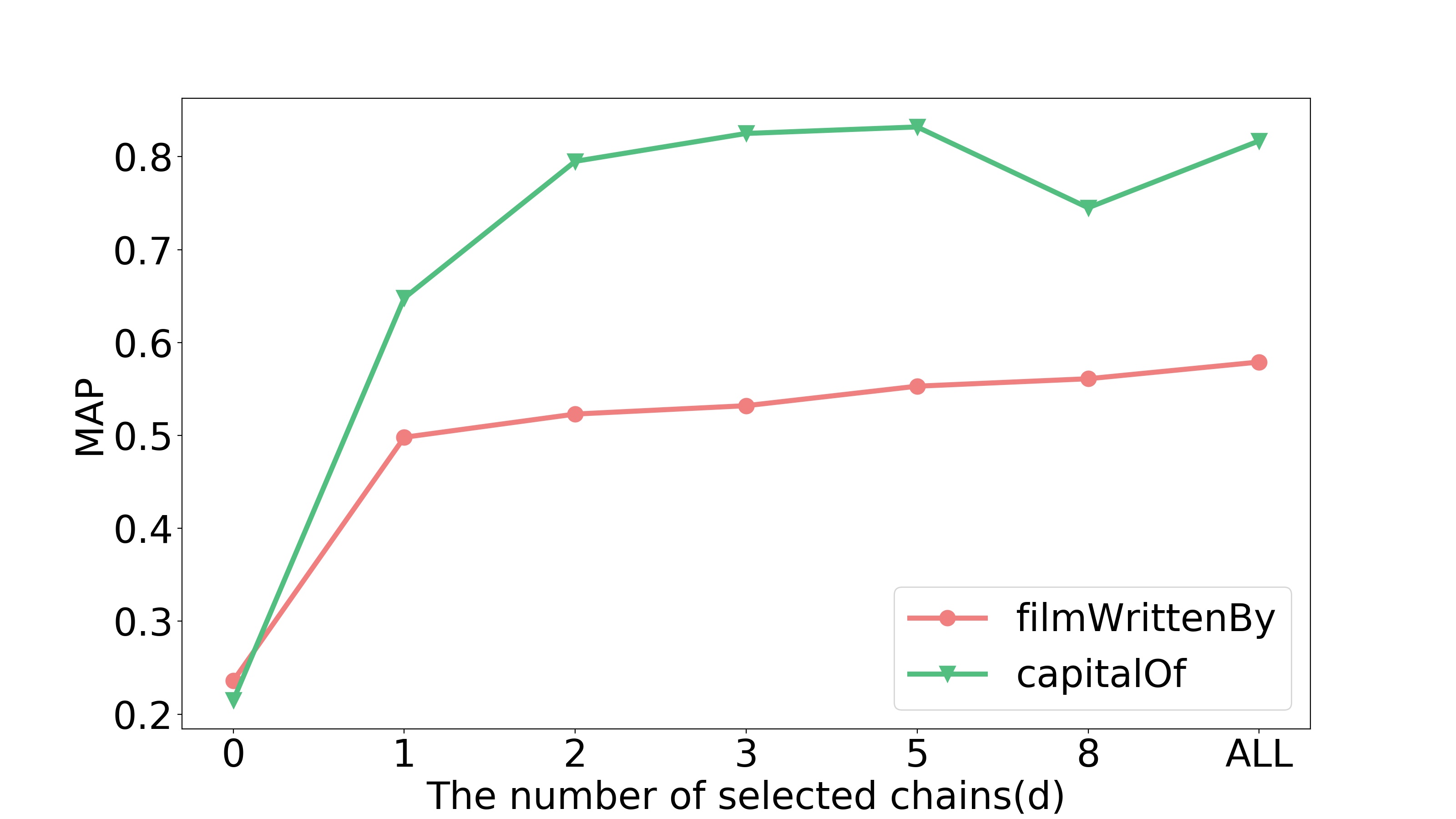}
\caption{MAP with $d$ increasing.\label{fig:curve}}
\end{figure}

\section{Additional Experiments on Top-$K$ Generation from the Single-Chain Baseline}
We add an additional experiment, \emph{Single-Chain Gen}, as an additional baseline in this part. Since we train the generator and predictor together at the same time in our method, we are interested in the performance of the predictor without knowing the target $d$ (i.e., the number of selected chains). In this experiment, we first train a singe-chain model to obtain a generator, then take the top $d$=2 or 5 chains from the  resultant generator and train the predictor separately. 
From the results shown in Table~\ref{tab:res_basemlp}, it can be observed that our proposed model also outperforms this new baseline. Hence our model does capture the conjunction information among the chains during the subset selection procedure in the generator phase.
{\footnotesize\begin{table*}[h]
\centering
\addtolength{\tabcolsep}{0.1pt}
{\footnotesize\begin{tabular}{c l c c c c c cc} 
\toprule
& \multirow{2}{*}{\bf Relation}&\bf  Single-Chain  &  \multicolumn{2}{c}{\bf Singe-Chain Gen} & \multicolumn{2}{c}{\bf Ours } & \multirow{2}{*}{\bf DeepPath} &\multirow{2}{*}{ \bf MINERVA}\\
&  &\bf Baseline & \bf $\bm d$=2& \bf $\bm d$=5 &\bf $\bm d$=2 &\bf $\bm d$=5 & &\\
\midrule
\multirow{11}{*}{\rotatebox{90}{\bf NELL-995}}
& athletePlaysForTeam & 0.872&0.898&0.913&0.940& \bf 0.947&0.750 &0.824\\
& athletePlaysInLeague &0.962 &0.957&0.977&0.977 & \bf 0.983&0.960& 0.970 \\
& athleteHomeStadium & 0.892 &0.859&0.856&\bf0.896&0.895&0.890& 0.895\\
& athletePlaysSport & 0.916&0.911&	0.978&0.978&0.982&0.957 &\bf0.985\\
& teamPlaySports & 0.728 &0.690&0.775&0.769 &0.782 &0.738&\bf0.846\\
& orgHeadquarterCity &{\bf0.957}&0.955&	0.953&0.932&0.907&0.790&0.946\\
& worksFor & 0.794 &0.859&	0.850&0.842 & \bf0.849&0.711 &0.825\\
& bornLocation &0.823&0.906&0.861&\bf 0.902&0.850&0.757&0.793\\
& personLeadsOrg & 0.833&0.817&	0.784&0.832&0.813& 0.795&\bf0.851\\
& orgHiredPerson & 0.833&0.833&	\bf0.852&0.825&0.814&0.742&0.851\\
\cline{2-9}
& \emph{Average} &\it 0.861&\it0.868&\it	0.880&\textbf{\emph{0.889}}&\it 0.882&\it 0.809&\it 0.879
\\
\midrule
\multirow{11}{*}{\rotatebox{90}{\bf FB15K-237}}
& teamSports &0.740&0.746&0.743&0.739&0.769&\bf0.955&-\\
& birthPlace	&0.463		&0.517	&0.512&0.505	&\bf 0.566	&0.531&-\\
& filmDirector&0.303&0.271&0.272&0.368&0.411&\bf 0.441&-\\
& filmWrittenBy 	&0.498	&0.523	&0.544&0.516	&\bf0.553	&0.457&-\\
& filmLanguage&0.632&0.687&0.684&0.665&\bf0.678&0.670&-\\
& tvLanguage&\bf0.975&0.967&0.968&0.962&0.957&0.969&-\\
& capitalOf &0.648&0.740&0.758&0.795&\bf0.825&0.783&-\\
& orgFounded &0.465&0.441&0.472&0.407&\bf0.490&0.309&-\\
& musicianOrigin & 0.376&0.419&0.468&0.408&\bf0.516&0.514& -\\
& personNationality &0.713&0.813&0.825&0.806&\bf 0.828&0.823&-\\
\cline{2-9}
& \emph{Average} &\it 0.581&\it 0.612	&\it 0.625&\it 0.617	&\textbf{\emph{0.659}}&\it 0.645&-\\

\bottomrule
\end{tabular}}
\caption{Overall Results (MAP) on NELL-995 and FB15K-237 single chain generator and MLP predictor.}
\label{tab:res_basemlp}
\end{table*}}

\end{document}